\documentclass[sigconf]{acmart}
\usepackage{subcaption}
\usepackage{algorithm2e}
\usepackage{algpseudocode}
\usepackage[table]{xcolor}

\setcopyright{acmlicensed}
\copyrightyear{2025}
\acmYear{2025}
\acmDOI{}
\acmISBN{}
\acmConference[]{CARS@RecSys '25}{Sep 22--26, 2025}{Prague, Czech Republic}
\acmBooktitle{the 19th ACM Conference on Recommender Systems (RecSys '25), September 22--26, 2025, Prague, Czech Republic}

\begin{document}
\title{FinTRec: Transformer Based Unified Contextual Ads Targeting and Personalization for Financial Applications}

\author{Dwipam Katariya}
\orcid{0009-0009-1058-1244}
\affiliation{%
    \institution{Capital One, AI Foundations}
  \city{McLean}
  \country{USA}
  }

\author{Snehita Varma}
\orcid{0009-0005-9559-743X}
\affiliation{%
\institution{Capital One, AI Foundations}
  \city{San Francisco}
  \country{USA}
    }

\author{Akshat Shreemali}
\orcid{0009-0003-4481-150X}
\affiliation{%
\institution{Capital One, AI Foundations}
  \city{New York}
  \country{USA}}

\author{Benjamin Wu}
\orcid{0000-0003-3874-7030}
\authornote{Contributed to the study when at Capital One. Now at NVIDIA}
\affiliation{%
\institution{Capital One, AI Foundations}
  \city{San Jose}
  \country{USA}}

\author{Kalanand Mishra}
\orcid{0000-0002-1832-1537}
\affiliation{%
\institution{Capital One, AI Foundations}
  \city{San Jose}
  \country{USA}}

\author{Pranab Mohanty}
\orcid{0000-0002-2224-8467}
\affiliation{%
\institution{Capital One, AI Foundations}
  \city{San Jose}
  \country{USA}}


\renewcommand\footnotetextcopyrightpermission[1]{%
  \footnotetext{
\vspace{0.5em} 
\hrule 
\noindent\\
Permission to make digital or hard copies of all or part of this work for personal or
classroom use is granted without fee provided that copies are not made or distributed
for profit or commercial advantage and that copies bear this notice and the full citation
on the first page. Copyrights for third-party components of this work must be honored.
For all other uses, contact the owner/author(s). \\
© 2025 Copyright held by the owner/author(s). \\
Correspondence to: Dwipam Katariya <dwipam.katariya@capitalone.com>
}
}
\renewcommand{\shortauthors}{Katariya et al.}


\begin{abstract}
Transformer-based architectures are widely adopted in sequential recommendation systems, yet their application in Financial Services (FS) presents distinct practical and modeling challenges for real-time recommendation. These include:a) long-range user interactions (implicit and explicit) spanning both digital and physical channels generating temporally heterogeneous context, b) the presence of multiple interrelated products require coordinated models to support varied ad placements and personalized feeds, while balancing competing business goals. We propose FinTRec, a transformer-based framework that addresses these challenges and its operational objectives in FS. While tree-based models have traditionally been preferred in FS due to their explainability and alignment with regulatory requirements, our study demonstrate that FinTRec offers a viable and effective shift toward transformer-based architectures. Through historic simulation and live A/B test correlations, we show FinTRec consistently outperforms the production-grade tree-based baseline. The unified architecture, when fine-tuned for product adaptation, enables cross-product signal sharing, reduces training cost and technical debt, while improving offline performance across all products. To our knowledge, this is the first comprehensive study of unified sequential recommendation modeling in FS that addresses both technical and business considerations.
\footnote{The modeling referenced in this research is not reflective of actual, granular activities or tasks unique to any specific organization and rather are common across the financial services industry}
\end{abstract}



\maketitle

\section{Introduction}
\begin{figure*}
    \centering
    \includegraphics[scale=0.48]{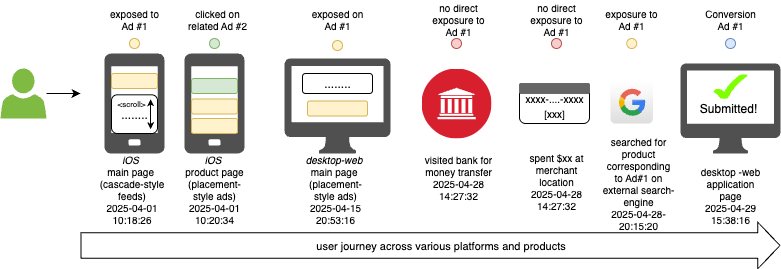}
    \caption{Illustrative example of user's journey across channels, products and pages}
    \Description[Illustrative example of user's journey across channels, products and pages]{}
    \label{fig:journey}
\end{figure*}

Financial Service (FS) enterprises maintain a rich history of user's multichannel interactions, with millions of users generating billions of interactions daily across several digital products and physical channels. Users transact with credit and debit cards, browse digital channels, talk to bank associates to accomplish tasks that may or may not directly signal product interests and even get exposed to the related ads on external search engine before an actual purchase, as illustrated in Figure \ref{fig:journey}. (See Appendix \ref{appendix:cj} for more details on customer journey). Sequential recommendation systems \cite{sasr, bert4rec, spotify, alibabem1, alibaba, seqgnn, genrec, trans4rec, gtransrec, sessionrec, cntxt-awr} have used explicit and implicit signals \cite{ren2018learning, Fan_2022, din, dien, dfn, dstn, temp4ctr} such as clicks, dislikes/likes, and others, combining longitudinal and heterogeneous contextual data. (See Appendix \ref{appendix:related work} for more on related work). However, prior work has not explored other dynamic and non-digital signals such as transaction and payments history, despite their dominance in FS. Representing such temporal multi-channel, multi-product behavior is important for accurate real-time personalization, specifically at scale under low-latency constraints for high throughput applications. This requires a principled design of sequential representations and optimization of architectural and embedding strategies \cite{pinnerformer, sum}. FS also has a varied content mix with various objectives, as shown in Table \ref{tab:item}. Hence, optimizing solely for Click-Through-Rate (CTR) risks adverse selection, promoting low-utility, clickbait items. Similarly, optimizing solely for Conversion-Rate (CVR) may degrade user experience. Furthermore, a mix of content is served at different touchpoints by interrelated products such as feed-style ranking and placement-style ads targeting. Since feed-style personalization may globally rank locally personalized ads content, it demands cross-product awareness. This proliferates product-specific models which are effective in isolation but result in increased maintenance cost and technical debt \cite{unicorn}. Moreover, such siloed models fail to leverage valuable cross-product interactions, leading to suboptimal performance \cite{unified}. Recent research \cite{unified, user, unifiedssr, kuaisar, ugsr} has explored the integration of search and recommendation systems, primarily within the domains of e-commerce, travel, and digital media. However, this work does not sufficiently address the FS sector, where large enterprises operate multiple interrelated products beyond search and recommendation systems. Moreover, FS still relies on tree-based approaches for all Machine Learning (ML) modeling tasks \cite{deeptlf, trxnclass, creditcardfraud}, primarily due to regulatory alignment (e.g., explaining loan application denials). This approach necessitates extensive feature engineering by domain experts, a process that is both time-consuming and complex. The evolving nature of these feature sets further complicates the management of large-scale ML systems, making it difficult to maintain and scale models efficiently. Hence, we present FinTRec, a unified transformer-based framework for personalized advertising and servicing needs for FS that: a) considers the user's raw sequential and heterogeneous context eliminating the need for feature engineering, b) balances both business value and user experience with a multi-objective function, c) addresses infrastructure complexity and cost in order to enable real-time inference by relying on a proprietary foundational model (FM) to capture the user's years worth of transactions, payments, statements and other context and, d) when fine-tuned (FT) for product adaptation, enables improved local optimization through cross-product awareness, reduces technical debt and deployment complexity.

\begin{table*}
    \caption{Content mix served on FS digital products}
    \label{tab:item}
    \begin{tabular}{lll} 
        \toprule
        Content Type & Illustrative Examples & Feedback \\
        \midrule
        Marketing & open new accounts - credit card, loan, bank etc.. & Convert \\
        Servicing & enroll in paperless statement, check savings balance etc.. & Click \\
        Third-party & 4x miles on Expedia, 5x cashback on Etsy, etc.. & Click $/$ Convert \\
         \bottomrule
    \end{tabular}
\end{table*}

\section{Task definition}
\textbf{Click-Through-Rate} as \textit{Next item Prediction}: Let $i \in \mathcal{I_{T}}$ represent the candidate item shown to a user $u \in \mathcal{U}$ at time $t \in \mathcal{T}$. Hence, $\hat{p}_{\text{CTR}}(i, u)_t = \mathbb{P}(\text{click}_{i} = 1 \mid I_{t-1}, u_t)$. \\
\textbf{Conversion-Rate}: Traditionally, conversion rate (CVR) is defined as the probability of conversion given a user clicks. However, in FS platforms, exposure alone, without a click, can influence conversions. Additionally, users may convert without any platform exposure, arriving via external sources like search engines or direct links. Hence, $p{\text{CVR}}(i, u)_t = \mathbb{P}(\text{convert} = 1\mid i, u_t)$.\\ \textbf{Final Ranking}: CTR reflects short-term engagement, while CVR captures long-term value. Optimizing solely for CTR can promote low-quality or clickbait content with limited business impact. In Platform Generated Content (PGC), where content supply is platform-driven, ranking must adapt quickly to shifting business priorities influenced by economic and policy changes. To support this, we introduce an urgency signal $us(i) \in \mathbb{R}_{\geq 0}$, quantifying the strategic importance of promoting item(i) at a given time(t) set at the campaign owner level. The final Ranking Score(RS) is computed as:
\begin{equation}
\text{RS}(i,u)_t =  \lambda_{us} \cdot us(i)_t + \lambda_{ctr} \cdot p{\text{CTR}}(i,u)_t + \lambda_{cvr} \cdot p{\text{CVR}}(i,u)_t \cdot v(i)_t
\end{equation}
where: $p{\text{CTR}}(i,u)_t$: predicted probability of a click after calibrating $\hat{p}_{\text{CTR}}(i, u)_t$, $p{\text{CVR}}(i,u)_t$: predicted probability of conversion, $v(i)$: present value(PV) per conversion in dollars, $us(i)$: urgency score, $\lambda_{us} \in \mathbb{R}_{\geq 0}$: stakeholder-controlled urgency weighting coefficient, $\lambda_{ctr} \in \mathbb{R}_{\geq 0}$: CTR weighting coefficient, $\lambda_{cvr} \in \mathbb{R}_{\geq 0}$: CVR weighting coefficient. This formulation enables flexible balancing of long-term value optimization and short-term business priorities, while preserving ranking interoperability. The urgency term acts as a soft override, allowing stakeholders to elevate the prominence of specific item categories without entirely bypassing user-preference signals (e.g., fed policy change affecting savings account APY). Since, PV is realized only when a conversion occurs, $p{\text{CVR}}(i,u)_t \cdot v(i)_t$ estimates the expected value per conversion. It should be noted that other signals do not have a monetary value associated with them.
\footnote{For brevity,  $\hat{p}_{\text{CTR}}(i, u)_t$ will be referred as pCTR and $p{\text{CVR}}(i,u)_t$ will be referred as pCVR}

\section{Methodology}
\subsection{Data Pre-processing}
\label{data:preprocess}

\begin{figure}[h!]
        \centering
        \includegraphics[scale=0.35]{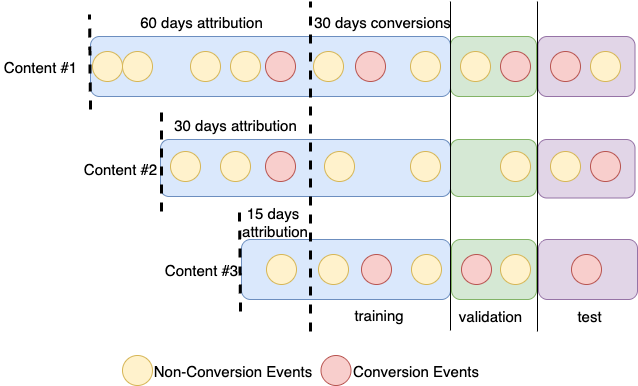}
        \caption{Yellow circle represent non$-$conversion events including impressions. Red circles represent conversions. For each conversion, item $\#1$ queries data back to $60$ days, while item $\#2$ and item $\#3$ queries data back to $30$ and $15$ days respectively. Then the data is temporally split for training/validation/test}
        \Description[Yellow circle represent non$-$conversion events including impressions. Red circles represent conversions. For each conversion, item $\#1$ queries data back to $60$ days, while item $\#2$ and item $\#3$ queries data back to $30$ and $15$ days respectively. Then the data is temporally split for training/validation/test]{}
        \label{fig:split}
\end{figure}

\begin{figure}[h!]
        \centering
        \includegraphics[scale=0.4]{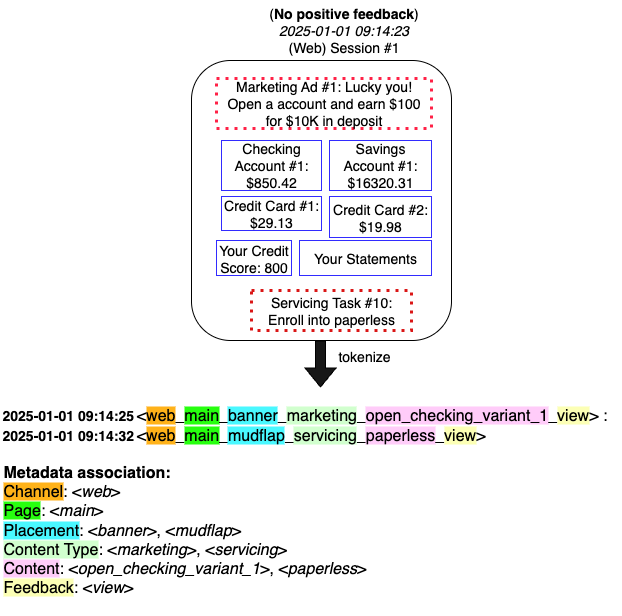}
        \caption{Illustrative example of intra-page tokenization. Tokens are grouped and arranged by their timestamps across pages}
        \Description[Illustrative example of intra-page tokenization. Tokens are grouped and arranged by their timestamps across pages]{}
        \label{fig:seq_token}
\end{figure}

Clickstream activities are stored at the most granular level of web and mobile activity, along with the timestamp. Individual transactions - such as payments, ATM usage, call center events - are also stored in a tabular format at the interaction level and serve as valuable context signals for online ad targeting and personalization \cite{timesync, dce}. A notable property of this data is its composition of two distinct signal types: dynamic and static ~\cite{timesync, fatatrans}. Dynamic signals, such as transactions, payments and money transfers, are characterized by their frequent changes over time. In contrast, Static signals, like a user's product ownership, maintain a persistent state over a longer temporal horizon. Table ~\ref{tab:signals} illustrates sample data sources used for this study. A fundamental requirement for an accurate real-time recommendation is low latency feature fetching and transformation. This is particularly challenging when leveraging an extensive history of user interactions, including offline touchpoints such as transactions and payments. Given the impracticality of performing these transformations in real-time, due to both the volume and heterogeneous dimensionality of the data, we rely on our proprietary FM to aggregate dynamic heterogeneous context to point-in-time representation.\\  
Our legacy Random Forest (RF) model \cite{deeptlf, trxnclass, creditcardfraud} processes expert-engineered features derived from the dynamic and static context and clickstream data (e.g., total web/mobile logins and product page visits in 7/14/30 days, avg. conversion rates in 7/14/30 days since fraud reported, utlization rate etc.). A critical limitation of the RF approach is its reliance on flattening user interaction histories into time-aligned snapshots per impression. This simplification overlooks fine-grained temporal dependencies and evolving user preference signals. Another drawback is the model's difficulty in handling high dimensional embeddings - such as a 768-dimensional FM embeddings - which can lead to unstable feature sampling and degraded performance. To address this, we applied PCA to reduce the embeddings to 32 dimensions before training. In contrast, FinTRec directly leverages raw user interaction histories and 768-dim FM embeddings. It segments these histories into sequences, each culminating in a conversion/no-conversion and/or click/no-click event, thereby rigorously preserving chronological order and temporal dependencies across sessions, channels and products. Due to content-specific regulatory requirements such as 30 days marketing-opt out notices, and varied conversion time window (e.g., enrolling in credit monitoring is a quicker conversion time frame compared to enrolling in a savings account) requires careful content-specific attribution process. Our data processing pipeline allows for flexible conversion attribution as illustrated in Figure \ref{fig:split} and filters user sequences based on marketing opt outs. \textit{Feature transformation}: Continuous context features undergo standardization. Categorical features like product enrollments, due to their multi-valued nature, are multi-hot encoded. Other categorical attributes (e.g., channel type, intra-page structured elements, layout, placements, device type) are tokenized as illustrated in Figure ~\ref{fig:seq_token}. FM embeddings are incorporated without further transformation. This process yields three fixed-dimensional context feature vectors - static context vector ($F_{s}$), dynamic context vector ($F_{d}$) , FM embedding vector ($F_{fm}$) per impression. For each user $u$, their tokenized interaction history is represented as a time-ordered sequence of items $S_u = (i_{t-l}, \dots, i_{t-0})$, where $l$ is the sequence length. Each item $i_t \in \mathcal{I}$ in $S_u$ is associated with its context vector $\{F_s, F_d, F_{fm}\} \subseteq F_c$.

\begin{table*}
    \centering
    \caption{Sample data sources used for this study}
    \label{tab:signals}
    \begin{tabular}{lll} 
        \toprule
        Data Source & Description & Type\\
        \midrule
        Clickstream & Impressions and clicks across pages, users, platforms and products. & Dynamic \\& Multiple impressions and clicks are grouped together per user session \\& spanning from login to logout/idle > 30 mins \\
        
        Heterogeneous Context & 768-dim batched sequential representation of last 3 years & Dynamic \\& transactions, payments, statements etc. with proprietary encoder-only FM\\
        Product Enrollment & Products that user own & Static \\
        Tenures & User relationship with all the products (in days) & Static \\
        \textbf{........} & \textbf{........} \\
         \bottomrule
    \end{tabular}
\end{table*}

\subsection{Model Architecture}

\begin{figure}[h!]
    \centering
\begin{subfigure}[b]{0.5\columnwidth}
    \includegraphics[scale=0.51]{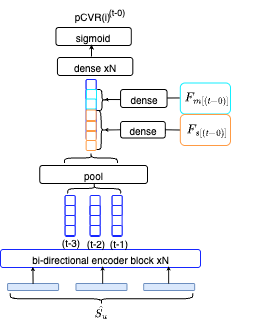}
    \caption{pCVR encoder-only architecture}
    \label{fig:encoder}
\end{subfigure}
\begin{subfigure}[b]{0.49\columnwidth}
    \includegraphics[scale=0.51]{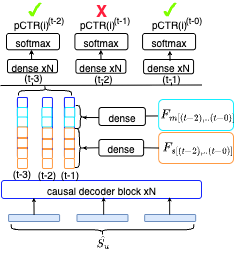}
    \caption{pCTR decoder-only architecture}
    \label{fig:decoder}
\end{subfigure}

\caption{The pCVR model employs an encoder-only architecture (Figure \ref{fig:encoder}), while the pCTR model utilizes a causal decoder-only architecture (Figure \ref{fig:decoder}). Only positive samples are used for training pCTR, hence (t-2) is skipped from final loss calculation}
\Description[The pCVR model employs an encoder-only architecture (Figure \ref{fig:encoder}), while the pCTR model utilizes a causal decoder-only architecture (Figure \ref{fig:decoder}). Only positive samples are used for training pCTR, hence (t-2) is skipped from final loss calculation]{}
\end{figure}

A decoder-only architecture is used for pCTR (Figure \ref{fig:decoder}), whereas an encoder-only architecture is used for pCVR (Figure \ref{fig:encoder}). Encoder architectures are widely adopted for modeling conversion behavior, while decoder architectures are commonly used for next click prediction, as the sequential dependency of user clicks and immediate feedback naturally aligns with the autoregressive properties of decoders. Although conversion modeling involves long-term attribution and delayed feedback, there is no theoretical limitation that precludes the use of decoders for conversion or encoders for click prediction. In practice, however, the prevailing choice is often influenced by practical considerations, such as data storage schemas and serving infrastructure, which tend to favor encoders for conversion modeling and decoders for click prediction.

\textbf{Sequence Processing}: For both pCTR and pCVR models, the pipeline first aggregates $F_d$ since user's most recent digital session into a dynamic context vector $F_{d(t)}$. $S_u$ tokens are mapped to dense vectors via an input embedding layer and fused with the resulting item token embeddings at (t) with $F_{d(t)}$ to form a single, combined embedding vector. Given that $F_{s}$ is time-invariant and $F_{fm}$ are inherently sequential, the pCVR model utilizes $F_{s}$ and $F_{fm}$ corresponding to the latest interaction ($t-0$). In contrast, the pCTR model retains $F_{s}$ and and $F_{fm}$ for all the time points within the sequence. \textbf{Order and Temporal Encoding}: To preserve the intra-session interaction order, we apply positional encodings $p_t$ as proposed by \cite{vaswani2023attentionneed}. To capture temporal dynamics, we construct a multivariate timestamp ${e_t}$ representation by decomposing timestamp into dayofweek, weekofmonth, hourofday etc. We then compute an element-wise dot product between the $p_t$ and ${e_t}$, and add the result to the input sequence embeddings resulting in $\hat{S_u}$. \textbf{Causal Decoder Block}: For pCTR task, We utilize a causal decoder-only architecture, where self-attention is masked based on timestamps rather than token positions, in contrast to the default formulation in \cite{vaswani2023attentionneed} resulting in: $\mathbf{h}_t = \text{Decoder}(\hat{S_u})$. This forces strict time ordered look-back constraint. \textbf{Bi-Directional Encoder Block}: For pCVR, We adopt a bi-directional encoder block \cite{vaswani2023attentionneed} resulting in $\mathbf{h}_u = \text{Encoder}(\hat{S_u})$. We then, pool the representations from $h_{(t-l)}$ to $h_{(t-0)}$ to capture a global summary of the interaction sequence: \textbf{Static and FM Context Fusion}: The decoder's output $\mathbf{h}_t$ is concatenated with [$f_{fm(t)}$, $f_{s(t)}$] and [$f_{fm(0)}$, $f_{s(0)}$] for encoder's output $\mathbf{h}_u $ . This combined representation is passed through a multi-layer perceptron (MLP) followed by Sigmoid (in case of pCTR) and Softmax (in case of pCVR). \textbf{Loss Functions}: \textit{${\text{pCVR}}$}: For training the Random Forest model with down-sampled negatives, we use the standard Binary Classification loss followed by item-specific calibration. Transformers is trained with BCE and does not require calibration. \textit{$\text{pCTR}$}: We adopt a modified Next Item Loss (as shown in eq.2) that includes only clicked items. Since certain item must be exposed regardless of clicks, the model is not optimized for predicting the immediate next item, but instead focuses on learning from positively engaged signals.
\begin{equation}
\text{Next Item Loss} = -\frac{1}{\sum_{u \in \mathcal{U}} |\mathcal{P}_u|} 
\sum_{u \in \mathcal{U}} \sum_{t \in \mathcal{P}_u} 
\-log \left( \frac{ \exp(\hat{y}_{u,t}[y_{u,t+1}]) }{ \sum_{j=1}^K \exp(\hat{y}_{u,t}[j]) } \right)
\end{equation}
where,  $\mathcal{U}$: set of users, $\mathcal{P}_u$: time steps t for user u such that a click occurred at t+1, $y_{u,t+1}$: the ground-truth clicked item at time t+1, $\hat{y}_{u,t} \in \mathbb{R}^K$: the logits (predicted scores) over the item vocabulary at time t

\subsection{Product Adaptation FT}
\label{section:FT}
First, we collapse customer interactions across all the existing products and align them on a single flat timeline \cite{timesync}, followed by pre-training a FinTRec-decoder-only model. We then adopt Low-Rank Adaptation (LoRA) \cite{lora-ft} for new product adaptation. During fine-tuning, only the low-rank product-specific matrices are updated. To incorporate product-specific semantics, we extend the pre-trained token embedding matrix with new learnable embeddings. Let $E \in \mathbb{R}^{V \times d}$ denote the original token embeddings for a vocabulary of size $V$ and hidden size $d$. We introduce $E_{\text{new}} \in \mathbb{R}^{V_{\text{new}} \times d}$, representing embeddings for $V_{\text{new}}$ new product(s). The combined embedding matrix becomes:\[
E' = \begin{bmatrix}
E \\
E_{\text{new}}
\end{bmatrix} \in \mathbb{R}^{(V + V_{\text{new}}) \times d}
\]
In our setup, $E$ remains frozen during fine-tuning, while $E_{\text{new}}$ is optimized jointly with the LoRA parameters. Formally, let $\theta_{\text{LoRA}}$ represent all LoRA parameters across selected transformer layers, and $E_{\text{new}}$ the trainable token embeddings. The model is fine-tuned by minimizing a task-specific loss $\mathcal{L}$ over labeled training data $(S_u, F_c, y)$:\[
\min_{\theta_{\text{LoRA}}, E_{\text{new}}} \; \mathcal{L}\left(f_{\text{base}}(S_u; F_c; \theta_{\text{frozen}}) + f_{\text{LoRA}}(S_u; F_c; \theta_{\text{LoRA}}, E_{\text{new}}), y\right)
\]
Here, $f_{\text{base}}$ is the frozen backbone model, and $f_{\text{LoRA}}$ represents the learned low-rank adaptations and new token embeddings. This formulation ensures that only a small subset of model parameters along with new tokens are updated, enabling scalable and efficient adaptation to new products.\\
\textbf{Linear Probing (LP)} enables lightweight fine-tuning by updating only the final dense layer, while keeping all other model parameters frozen. 
\textbf{Full Fine-Tuning (F-FT)} All the model parameters of the base transformer layers are updated.\\
Token embedding extension is also applied to both LP-FT and F-FT. Since different products have different output dimensions, a new output adapter is added by replacing the base model head. Output adapter is replaced for all three FT strategies. For F-FT and LoRA-FT, we also unfreeze parameters in the dense layer that correspond to $F_s$ and $F_{fm}$. The goal of LP-FT was to set a lower bound on the maximum performance gain we could achieve with minimal updates; hence, the dense layers corresponding to $F_s$ and $F_{ffm}$ were kept frozen.

\begin{figure}[h!]
    \centering
    \includegraphics[scale=0.6]{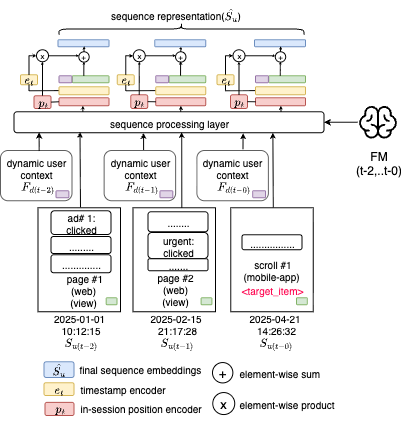}
    \caption{Dynamic context ($F_d$) and Clickstream sequence ($S_u$) are processed to form sequential user representations ($\hat{S_u}$). Static ($F_s$) and FM embeddings ($F_{fm}$) are time-aligned with $\hat{S_u}$}
    \Description[Dynamic context ($F_d$) and Clickstream sequence ($S_u$) are processed to form sequential user representations ($\hat{S_u}$). Static ($F_s$) and FM embeddings ($F_{fm}$) are time-aligned with $\hat{S_u}$]{}
    \label{fig:sequence_proc}
\end{figure}

\subsection{Experimental Setup}

We use various data sources from the internal data ecosystem as mentioned in Table ~\ref{tab:item}. For model training, we apply a temporal split of 90 days for training, 7 days for validation, and 7 days for testing on clickstream data. We focus our study on 30M+ user's sequences, 200+ items, 1BN+ interactions, and 10+ data sources across mobile and web channels. \\ 
\textbf{Metrics}: We adopt widely used evaluation metrics: log loss (for \textit{${\text{pCVR}}$}) and ranking recall@topk (for \textit{${\text{pCTR}}$}). \textit{Log Loss}: quantifies the uncertainty in predictions required for trustworthy monetary value assessments. \textit{Ranking Recall}: 99\% of sessions display a maximum of five items (across products) at any given time. An effective personalization system aims to elevate relevant content to the highest possible rank within this limited display. Ranking Recall directly quantifies this performance, measuring how effectively relevant items are surfaced in such constrained visual interfaces. Recall@topk is given as: 
\begin{equation}
\text{Recall@top-k} = \frac{1}{\sum_{u \in \mathcal{U}} |\mathcal{P}_u|} \sum_{u \in \mathcal{U}} \sum_{t \in \mathcal{P}_u} \mathbb{1} \{ I_{u,t+1} \in \hat{R}_{u,t}^n \}
\label{eq1}
\end{equation}
as such, $\hat{R}_{u,t}^n$ is the top-k recommended items at time t for user u.\\
\textbf{Product Adaptation}: The effectiveness of our product adaptation approach was evaluated using a leave-product-out validation scheme. Specifically, product-specific interactions were held out during the pre-training phase before being subjected to the dedicated product adaptation strategy outlined in Section ~\ref{section:FT}. We experiment our framework with both feed-style and placement-style products. We further tightly align our training with in-production inference in order to compute model performance metrics. For example, the PGC Marketing and Servicing model is inferred at the page level, while mobile homepage is inferred at each login. \\ \\
\textbf{Training Specifications}: The FinTRec model was implemented using the PyTorch framework. We utilized NVIDIA 8 A10G GPUs and 512 GB of RAM. We adopted Distributed Data Processing (ddp) from PyTorch Lightning for parallelization. Each experiment ran for 50 epochs using the AdamW optimizer with a learning rate $1 x 10^-4$, a weight decay of $1 x 10^-5$ and a batch size of 64 per epoch. Model with the lowest validation log loss was selected. We did not employ early stopping. We utilize 10 multi-head attention blocks with hidden dimensions of 256 and input/output embedding dimension of 512.

\definecolor{lightgreen}{HTML}{CCFFCC}
\section{Offline Results}
\begin{table}[h!]
\centering
  \caption{Experimental Results of FinTRec for pCVR task on PGC marketing content. Best results are boldfaced and highlighted}
  \label{tab:results}
  \begin{tabular}{lcc}
    \toprule
    Model&Log Loss \\
    \midrule
    RF + Feature Set & 0.0984  \\
    RF + Feature Set (In Prod) & 0.0938 \\ 
    \hspace{0.12cm} + FM Embeddings \\
    \rowcolor{lightgreen} FinTRec & \textbf{0.0439} \\
    \hspace{0.12cm} w/o time embeddings & 0.0481\\ 
    \hspace{0.12cm} w/o FM embeddings & 0.0605 \\
    \hspace{0.12cm} context window=1 & 0.1135 \\
    \hspace{0.12cm} context window=5 & 0.0844 \\
    \hspace{0.12cm} context window=50 & 0.0478 \\
    \hspace{0.12cm} context window=120 & \textbf{0.0439} \\ 
    \bottomrule
\end{tabular}
\end{table}

\begin{table}[h!]
\caption{Experimental Results of FinTRec when FT for product adaptation with pCTR objective. Recall@1,5 shows relative lift in (\%) over the product-specific baseline. Best results are boldfaced and highlighted. Second best are underlined.}
\label{tab:ft-results}
\begin{tabular}{llcc}
    \toprule
    Product&Model&Recall@1&Recall@5\\
    \midrule
    PGC & (Product-Specific) & 0.00 & 0.00 \\
     Servicing & FinTRec & -5.24 & +2.90 \\
     (placement-style) & F-FT & \cellcolor{lightgreen} \textbf{+26.85} & \underline{+20.09} \\
     & LP-FT & +11.41 & +3.62 \\
     & LoRA-FT & \underline{+24.21} & \cellcolor{lightgreen} \textbf{+20.11} \\
    Mobile Homepage & (Product-Specific) & 0.00 & 0.00 \\
    (feeds-style) & FinTRec & -8.64 & -16.34 \\
    & F-FT & \underline{+13.85} & \underline{+5.15} \\
    & LP-FT& +5.36 & +1.16 \\
    & LoRA-FT & \cellcolor{lightgreen} \textbf{+14.11} & \cellcolor{lightgreen} \textbf{+5.16} \\
    Third-Party & (Product-Specific) & 0.00 & 0.00 \\
    Marketing & FinTRec & -32.44 & -24.14 \\
    (placement-style)& F-FT & \cellcolor{lightgreen} \textbf{+25.63} & \cellcolor{lightgreen} \textbf{+17.28} \\
    & LP-FT & +8.17 & +3.52  \\
    & LoRA-FT & \underline{+23.11} & \underline{+15.75} \\
  \bottomrule
        \hline
        \hline
        \multicolumn{4}{|p{1.0\columnwidth}|}{\raggedright \textit{LoRA-FT and LP-FT, required <5\% and <1\% of the pretrained model's parameters and <10\% and <5\% of its training time}} \\
        \hline
\end{tabular}
\end{table}

Table ~\ref{tab:results} reports FinTRec’s performance on the pCVR model for PGC marketing content. FinTRec (0.0439) substantially outperforms the production-grade RF baseline (0.0984) and RF + FM embeddings (0.0938). Removing temporal encodings (0.0481) or FM embeddings (0.0605) degrades performance, demonstrating the importance of temporal dynamics and financial context. Long-range dependencies are prominent in FS, as demonstrated by the increased accuracy with a larger context window, a factor often overlooked or subdued by expert-engineered features. \textit{\textbf{Product Adaptation}}: Table ~\ref{tab:ft-results} presents pCTR model gains across PGC Servicing, Mobile Homepage, and Third-Party Marketing. The Product-Specific model is a model developed in isolation based on the current framework, and incorporates both static and dynamic signals, as well as FM Embeddings. F-FT achieves the highest lift (up to +26.85\% Recall@1). LoRA-FT reaches comparable performance (+24.21\% Recall@1) with <5\% parameter updates, and LP-FT offers smaller gains with <1\% parameter fine-tuning, underscoring the importance of full temporal adaptation. LoRA-FT strikes a balance between model performance and training cost by improving code base standardization. FinTRec without fine-tuning under performs in most cases, due to: a) differences in pCTR across products , specifically negatively affecting Third-Party more than PGC Servicing, where pCTR for PGC Servicing is marginally higher and b) a lack of product-specific temporal alignment.

\section{Visit Ablation} 
Understanding the contribution of individual user touch points and their interaction effects within a model is critical for multiple reasons. Beyond its utility for budget attribution, this interpretability is essential for regulatory compliance, such as with Fair Lending laws, the Fair Housing Act, and Model Risk Management. It also helps pave the way for consumer-facing explainability, as is increasingly mandated by regulatory bodies like the Consumer Financial Protection Bureau (CFPB) and the European Union's General Data Protection Regulation (GDPR) via the "right to explanation". Unlike aggregate feature importance, visit-level attribution makes such explanations possible, strengthens consumer trust and provides temporal granularity, enabling compliance teams and regulators to trace why a particular sequence of behaviors led to a specific recommendation. Hence, we employed multiple attention and gradient based explainability frameworks like attention weights \cite{exp1} and GRAD-SAM \cite{exp-gradsam}. Attention weights provide direct interpretability as the Transformer architecture learns how much weight to assign to each visit. Specifically we extract the attention matrix and average across heads and layers to obtain a scalar weight per visit, then normalize the scores to create a visit importance distribution. While Grad-SAM combines both the self attention weights and their gradients, producing a single relevance map. Just a few touch points can possess predictive power comparable to that of the entire interaction sequence, as shown in Table \ref{tab:vist}. This finding highlights the efficacy of our explainable insight in an event of regulatory findings. Applicability and differences from various other post-hoc gradient and SHAP based methods like Integrated Gradients \cite{exp-ig} and Gradient-SHAP \cite{exp-gradshap} are yet to be studied on our data.

\section{Inference and Serving}
Real-time model deployment involves multiple execution stages to deliver final recommendations. We maintain a strict end-to-end latency requirement of 120ms at the 99th percentile for under 1500 api requests per second, irrespective of the number of model calls. Our solution is deployed on internally developed real-time serving platforms that handle both feature and model serving. The framework utilizes a combination of FM Embeddings and other contextual features computed and batched nightly, as well as features generated since the last batch job. At the time of a request, these features are retrieved and combined with a set of content determined to be eligible for the user, which is then provided to the model to generate the final ranking.

\section{Online A/B Experiments}
\begin{table}
  \caption{Log Loss vs A/B test performance. Log Loss and PV shows relative lift in \%. Act. PV shows actual realized PV in production. Est. PV when simulated  offline pre-deployment is calculated based on final RS. Feature set corresponds to our proprietary user context features set as illustrated in Table ~\ref{tab:signals}. Our simulation do not guarantee the magnitude of log loss improvement resulting in equal magnitude of PV improvement.}
  \label{tab:abtest}
  \begin{tabular}{llcc}
    \toprule
    Control&Test&Log Loss(\%) &Act. PV(\%) \\
    & & & (Est. PV(\%)) \\
    \midrule
    RF & RF & -7.00 & +3.75\\
    & + Feature Set &  & (<+6.08) \\
    
    RF & RF & -4.67 & +10.00\\
     + Features Set & + Features Set & & (<+37.50) \\ &+ FM Embeddings \\
    RF & FinTRec & -55.38 & TBD\\
     + Features Set  & + Features Set & & (<+41.50) \\ &+ FM Embeddings & \\
  \bottomrule
\end{tabular}
\end{table}

\begin{table}
  \caption{Visit quantification for explainability. AUROC is obtained from a model trained with important touch points. Relative AUROC in \% is reported compared to the best performing model from the Table ~\ref{tab:results}. (\%) is rounded to the nearest integer.}
  \label{tab:vist}
  \begin{tabular}{cc}
    \toprule
    Visit Importance&AUROC(\%) \\
    \midrule
    Most Important & -4.00\% \\
    Second Most Important & -5.00\% \\
    First Two Important Visit & -2.00\% \\
  \bottomrule
  \hline
        \hline
        \multicolumn{2}{|p{1.0\columnwidth}|}{\raggedright \textit{Visit importance identified with the Attention Tracing methods described in \cite {exp1, exp-gradsam} and averaged across both the methods }} \\
        \hline
\end{tabular}
\end{table}

\begin{figure}[h!]
        \centering
        \includegraphics[scale=0.32]{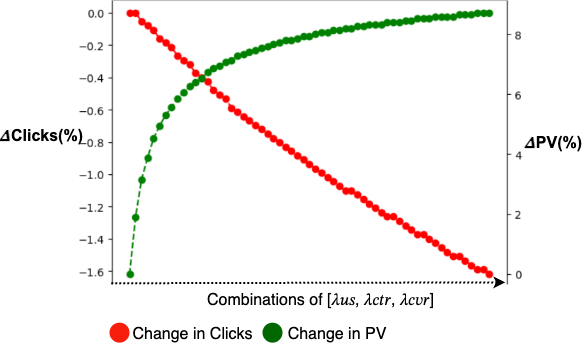}
        \caption{Change in expected clicks and PV based on offline simulation runs.}
        \Description[Change in expected clicks and PV based on offline simulation runs.]{}
        \label{fig:sim}
\end{figure}

We conduct offline simulations to assess a model's potential impact on downstream business outcomes before it is deployed to production. Our simulation methodology, inspired by established approaches from Netflix \cite{10.1145/3394486.3403229} and Yahoo \cite{10.1145/1935826.1935878}, systematically explores the search space of weighting coefficients ($\lambda_{us}, \lambda_{ctr}, \lambda_{cvr}$) to determine an optimal balance for the ranking score defined in Eq. 1. This process generates a sensitivity curve (Figure \ref{fig:sim}) that illustrates the trade-off between estimated Clicks and PV. The final coefficients are selected by product owners to align with current business objectives, typically by setting the values to match historically observed CTR and then noting the resulting increase in PV. Table \ref{tab:abtest} compares log loss reductions(\%) with both actual and estimated total PV lifts from live A/B tests(\%) and offline simulations. Using the latest dataset, we compare the top-ranked items from the current and new models using our final ranking objective, and estimate the lift in total PV. This evaluation shows that lower log loss generally correlates with higher user engagement, though the relationship is not strictly proportional. Adding the relevant feature set to the RF, the baseline yields a -7.00\% log loss reduction and a +3.75\% realized PV gain with offline estimate showing up to +6.08\% PV gain. Further incorporating FM embeddings improves PV by +10.00\%, with simulations estimating up to +37.50\% lift. The largest log loss reduction (-55.38\%) is achieved with FinTRec, with a projected simulated PV gain up to +41.50\%. While actual PV impact is pending, we conclude the offline results support FinTRec deployment. Furthermore, we also monitor other guardrail metrics like Fraud Rate (eg: \% Frauds reported in 30 days), 2hr. Call Rate post a user session, Credit Card On-Time Payments which may affect regulatory findings. In addition to guardrail metrics, tracking stakeholder-defined metrics is also essential to account for variations in PV, urgency scores and external factors like interest rates. For instance, business can decide thresholds for impression rates on specific content categories. This allows to detect unintended - a) model consequences early b) cannibalization due to content relatedness.

\section{Conclusion}
We introduces FinTRec and show superior performance over traditional and widely used tree-based baseline. We further show the effectiveness of incorporating time, long-range sequences, and implicit feedback via FM embeddings from years of transactional and other offline context. FM Embeddings enable real-time inference by eliminating the need for costly history retrieval. Fine-tuning across products further enhances performance through cross-product knowledge sharing, reduces infrastructure overhead, and improves codebase standardization. Historical A/B tests, simulations and explainability study confirms FinTRec's production readiness. \textbf{Limitations and Future Work:} a) pCTR and pCVR models are developed in separate codebases. Hence, future work will explore scalable architectures and unified frameworks for click and conversion models to further mitigate technical debt, maintenance, and training costs b) Our current FM embeddings are nightly batched and therefore stale, necessitating solutions for same day interaction awareness without incurring additional latency c) Although we applied explainability methods for visit importance in budget allocation and initial regulatory analysis, use-case-specific frameworks for finer alignment are yet to be fully explored. \textit{Broader Impact Statement:} While studied in context of Financial Services and Banking, this work is also applicable to diverse industries such as e-commerce, retail, travel, and media. We invite other researchers to expand our study further.

\section{Author Bio}
Dwipam Katariya(Mclean, VA), Snehita Varma(San Francisco, CA) and Akshat Shreemali(New York, NY) are Data Science Manager at Capital One. Kalanand Mishra(San Jose, CA) is Director of Data Science at Capital One. Pranab Mohanty(Seattle, WA) is Sr. Director of Applied Research at Capital One. They all are part of recommendation and personalization team within the AI Foundations org. Benjamin Wu worked as Data Science Manager at Capital One, before joining NVIDIA as Sr. Solutions Architect.

\begin{acks}
We thank Aditya Sasanur and Neil Kumar for their contribution to ML experiments. We thank Jesse Zymet for a thought provoking discussion and insights for highlighting value of the implicit feedback data. We are thankful to our collaborators from Foundation Models team. We thank Giri Iyengar for the support and sponsoring the study.
\end{acks}

\bibliographystyle{acm}
\bibliography{biblio}

\appendix
\section{Related Work}
\label{appendix:related work}

Recent advancements in recommendation systems have shown the importance of using sequential machine learning methods for representing a user's Point-of-Interest (POI) based on their past interactions and automatically recommending interested items. Users provide explicit and implicit feedback from interactions such as clicks, conversions, likes, dislikes, comments, shares, and/or dwell time. However, utilizing all user historical information poses challenges in capturing evolving user preferences over time. This has led to the emergence of session-aware recommendation (SAR) \cite{ligraph} models that adapt dynamically to ongoing user behavior. SAR has witnessed widespread adoption within academia as well as the industry, growing over 2x YoY since FY 2019 \cite{ligraph}. For instance, SAR has been widely studied within the e-commerce, travel, and entertainment industries. Recently, transformers have shown superior performance for SAR over RNNs and other traditional Machine Learning Models \cite{alibaba, kuaiformer, sasr, bert4rec, pinnerformer}. Although a plethora of research addresses peculiarities of transformers for SAR \cite{sasr, bert4rec}, these academic efforts often fail to address certain industrial challenges, which has limited their effectiveness in driving success in large-scale recommendation systems \cite{kuaiformer}. For example, AliBaba \cite{alibaba} introduced a behavior sequence transformer for e-commerce recommendation and demonstrated the superiority of sequential actions over an embedding and MLP paradigm—where raw features are embedded into low-dimensional vectors and passed into an MLP for final recommendation. However, while the model showed improvements, the study did not sufficiently address industrial challenges like scalability and inference latency for large-scale recommendation systems. Pinterest studied the effectiveness of deploying a model that leverages a user's historical actions within a large-scale recommendation system, focusing on an industrial solution capable of handling billions of pins (items) and 400M+ MAU. Their work highlighted the complexities of SAR, including using streaming infrastructure to capture the latest user activity and managing data to encode the user state in real-time \cite{pinnerformer}. Furthermore, the retrieval task—a precursor to the final recommendation—is responsible for selecting thousands of candidate items from billions of options. Applying Transformers directly to this large-scale operation in real-time is computationally prohibitive \cite{kuaiformer}. To address these challenges, a specialized architecture, KuaiFormer, was introduced by Kuaishou. Besides SAR studied in the e-commerce, travel, and entertainment domains, SAR has also been deployed in wealth management and, more recently, has started to gain adoption in retail banking. Several competitions on Kaggle held by Santander \cite{rnnforsar} highlight the growing need for recommending relevant retail banking products to users. Although these studies have been discussed in the realm of model architectures, datasets, evaluation, and bias detection, there are limited studies available on industrial applications of transformers in the Financial Services domain. And unlike digital-only platforms, users in FS also interact across various channels, for example, by logging into mobile and/or web applications, using their credit/debit cards, physical locations, etc., generating a rich history of user's sequential interaction \cite{timesync}. TIMeSynC \cite{timesync} addressed this through a data processing model architecture, yet failed to address the real-time aspect of data retrieval. Customer interactions across these channels are not only explicit but also leave trails of their implicit feedback. While several studies have focused on modeling positive feedback—such as clicks or conversions using sequential models—negative signals such as skips and exposure without engagement have been often overlooked \cite{Fan_2022}. DIN \cite{din} made an early attempt to capture the diversity of user interests by designing an adaptive activation mechanism, allowing the model to selectively attend to relevant historical behaviors based on current context. However, DIN is limited in its ability to model the evolution of latent interests over time \cite{dien}. DIEN \cite{dien} extended DIN by adding a latent temporal interest extractor and an explicit user behavior extractor, followed by a GRU-based sequential layer to learn dynamic user representations. Despite these improvements, DIEN still does not explicitly account for negative feedback signals. DSTN \cite{dstn} and DFN \cite{dfn} highlighted the importance of modeling both a user's positive and negative (exposure, unclicked) feedback for CTR prediction tasks. Exposure data is abundant, and recent studies demonstrated the usefulness of using such data in behavioral sequential tasks. For instance, XDM \cite{xdm} demonstrated the effectiveness of exposure data to help guide positive explicit clicks. DFN \cite{dfn}, DSTN \cite{dstn}, DIEN \cite{dien}, and XDM \cite{xdm} all incorporated users’ positive and exposure feedback independently, ignoring mutual interplay and contextual dependencies between them. Moreover, they also overlooked the influence of spatial page context on user behavior. RACP \cite{Fan_2022} addressed this gap by jointly modeling both the exposure and click sequences, capturing their interactions. RACP used a hierarchical attention architecture enabling the model to learn both intra-page context (how items on the same page influence each other) and inter-page interest shifts across sessions. Despite notable advancements, all these studies overlook the effect of sequence truncation in multi-feedback settings, undermining the cross-correlation between different feedback signals. TEM4CTR \cite{temp4ctr} employs a three-stage framework by first searching unclicked records close to clicked records, followed by an attention mechanism to extract relevant exposure information, and finally an interest extraction module to jointly extract a user’s latent interest representations from both the clicked and unclicked sequences for the CTR prediction task. However, these studies do not address other implicit feedback behavior data generated from cross-channel behaviors that are usually observed and dominant in the FS domain. Recent studies have shown the importance of combining cross-channel behavior for real-time intelligent personalization. \cite{alibabem1, albabaem2, youtube, jd, airbnb, googleplay, spotify} studied pre-trained transfer learning approaches and demonstrated the efficacy of embedding data streams combined with the final model to be suitable for real-world systems. However, FS has several such data streams, and combining them for low latency inference is yet to be studied. This requires combining pre-trained embeddings with sequential data. Tab-Transformer \cite{tabtransformer} contextualized categorical features but lacked contextualization across other feature types \cite{dlfortab}. Both FT-Transformer and Tab-Transformer studied the architectural changes on several datasets including bank marketing datasets; however, they lacked interaction signals from other heterogeneous data such as payments, transactions, etc., that are dominant in FS \cite{dlfortab}. Due to various data sources and multi-modality, developing embeddings on individual data streams and using them in downstream models has been previously discussed \cite{dce, tranformerfortrxn, pinnerformer, unittab}. \cite{dce} discussed clickstream embeddings as an input for fraud, intent prediction, and tasks in FS. Similarly, \cite{unittab} demonstrated the impact of payments and transaction embeddings on Fraud as a downstream task. Thus, these studies focused on single data stream representation and lacked combining representation from different data streams for accurate marketing and ads recommendation in FS. Furthermore, it’s a common practice to perform feature engineering and convert raw sequences into tabular data \cite{deeptlf, trxnclass} for machine learning algorithms such as Gradient Boosted Trees, more specifically in the FS and banking domain \cite{creditcardfraud} due to their higher explainability for closer regulatory alignment. However, this requires extensive feature engineering on longitudinal tabular data leading to long hours spent in developing accurate features. Traditionally, organizations have developed product-specific machine learning models to deliver personalized user experiences. While effective in isolation, this approach often results in fragmented infrastructure and substantial technical debt \cite{unicorn}. Moreover, such siloed models fail to leverage valuable cross-product interactions, leading to suboptimal performance \cite{unified}. Recent research \cite{unified, user, unifiedssr, kuaisar, ugsr} has explored unifying search and recommendation systems, primarily within the domains of e-commerce, travel, and digital media. However, this body of work does not sufficiently address the financial services (FS) sector, where enterprises often operate multiple interrelated products beyond just search and recommendation systems. In the FS industry, personalization efforts and advertising strategies have similarly relied on distinct, product-specific models, thereby overlooking potential gains from shared signals across products. Notably, naive attempts to merge signals from diverse products have revealed a fundamental trade-off: improvements in one task frequently degrade performance in others \cite{unified}. This challenge is compounded by the heterogeneity of sequential signals across products—each product may depend on distinct features, with only partial overlaps—resulting in redundant feature engineering and inefficient inference pipelines. Hence, in this study, we address all the gaps above and propose FinTRec tailored to unifying personalization and ads targeting in FS.

\section{Customer Journey}
\label{appendix:cj}
Users interact with FS through multiple channels and products to complete their tasks. As they navigate digital platforms, they are exposed to personalized ads and service messages, delivered via mechanisms such as placement-based targeting or feed-style personalization. However, digital exposure is only part of the journey—users also engage through offline touch-points, including visits to physical locations or interactions with call center agents. An illustrative user journey Figure ~\ref{fig:journey} shows seven touch-points over a month: four via digital platforms, one physical location, one transaction, and one external source (e.g., Google Search). High-value outcomes, such as product acquisitions, are not always attributable to digital ads alone but may result from this broader interaction landscape. It further gets complicated by FS’s diverse and interrelated product offerings (Table ~\ref{tab:item}), necessitating accurate capture of contextual information such as channel, page type, placement, and layout. Channels include email, SMS, mobile, and web applications. Ads can appear across hundreds of pages—e.g., the homepage, credit card pages, or other product-specific sections—each with multiple placements (e.g., welcome modules, account details, recent transactions, action buttons, or marketing messages). Interaction data is stored at a granular level across disparate databases, using various schema and formats. Each event is timestamped, enabling alignment along a user’s timeline for sequential modeling tasks. 

\end{document}